\documentclass[a4paper, 10pt, conference]{ieeeconf}

\newif\ifarwfinalcopy

\arwfinalcopytrue  

\IEEEoverridecommandlockouts                       

\usepackage{graphics} 
\usepackage{epsfig} 
\usepackage{mathptmx} 
\usepackage{times} 
\usepackage{amsmath} 
\usepackage{amssymb}  

\usepackage{lineno}
\usepackage{tikzpagenodes}
\usepackage{background}

\usepackage{url}
\usepackage{multirow}
\usepackage{xcolor}
\usepackage{listings}
\usepackage{siunitx}
\usepackage{graphicx}
\usepackage{subcaption}

\ifarwfinalcopy
\backgroundsetup{color=white}
\else

\linenumbers

\newcommand{\MyARWConfidentialLogo}{
\begin{tikzpicture}[remember picture,overlay]
\node[align=center,text=blue] at ([yshift=1em]current page text area.north) {\Large \#\#\# ARW 2025 SUBMISSION: CONFIDENTIAL REVIEW COPY \#\#\#};
\end{tikzpicture}%
}

\SetBgContents{\MyARWConfidentialLogo}
\SetBgPosition{current page.north west}
\SetBgOpacity{0.5}
\SetBgAngle{0.0}
\SetBgScale{1.0}

\fi

\title{LLM-Empowered 
Embodied Agent for Memory-Augmented Task Planning in Household Robotics}

\author{Marc Glocker$^{1,2}$, Peter Hönig$^{1}$, Matthias Hirschmanner$^{1}$, and Markus Vincze$^{1}$
\thanks{$^{1}$ Automation and Control Institute, Faculty of Electrical Engineering, TU Wien, 1040 Vienna, Austria {\tt\small \{hoenig, hirschmanner, vincze\}@acin.ac.tuwien.at}}%
\thanks{$^{2}$ AIT Austrian Institute of Technology GmbH, Center for Vision, Automation and Control, 1210 Vienna, Austria {\tt\small marc.glocker@ait.ac.at}}%
}

\begin{document}

\bstctlcite{IEEEexample:BSTcontrol}

\maketitle

\begin{abstract}
We present an embodied robotic system with an LLM-driven agent-orchestration architecture for autonomous household object management.
The system integrates memory-augmented task planning, enabling robots to execute high-level user commands while tracking past actions.
It employs three specialized agents: a routing agent, a task planning agent, and a knowledge base agent, each powered by task-specific LLMs. 
By leveraging in-context learning, our system avoids the need for explicit model training.
RAG enables the system to retrieve context from past interactions, enhancing long-term object tracking.
A combination of Grounded SAM and LLaMa3.2-Vision provides robust object detection, facilitating semantic scene understanding for task planning.
Evaluation across three household scenarios demonstrates high task planning accuracy and an improvement in memory recall due to RAG.
Specifically, Qwen2.5 yields best performance for specialized agents, while LLaMA3.1 excels in routing tasks.
The source code is available at: \url{https://github.com/marc1198/chat-hsr}

\end{abstract}

\begin{keywords}
Embodied AI, Task Planning, Memory Retrieval
\end{keywords}

\section{Introduction}
\label{sec:1_introduction}
Despite recent progress in robotics and artificial intelligence, robots still struggle to adapt flexibly to the diverse, dynamic situations of real-world environments, particularly in household settings~\cite{tellex_robots_2020}. While symbolic task planning with languages like the Planning Domain Definition Language (PDDL)~\cite{mcdermott_1998_2000} is effective in domains with fixed rules and predictable object categories, it lacks the adaptability required for open-ended household environments. In such settings, robots must deal with ambiguous user commands, detect novel or unstructured objects, and respond to constantly changing spatial configurations~\cite{tellex_robots_2020}. These limitations motivate our hypothesis that a modular LLM-driven system can enhance flexibility by leveraging natural language understanding, contextual reasoning, and memory-based adaptation. We provide a proof-of-concept implementation and assess its performance in real-world household tasks.

\begin{figure}
    \centering
    \includegraphics[width=1.0\linewidth]{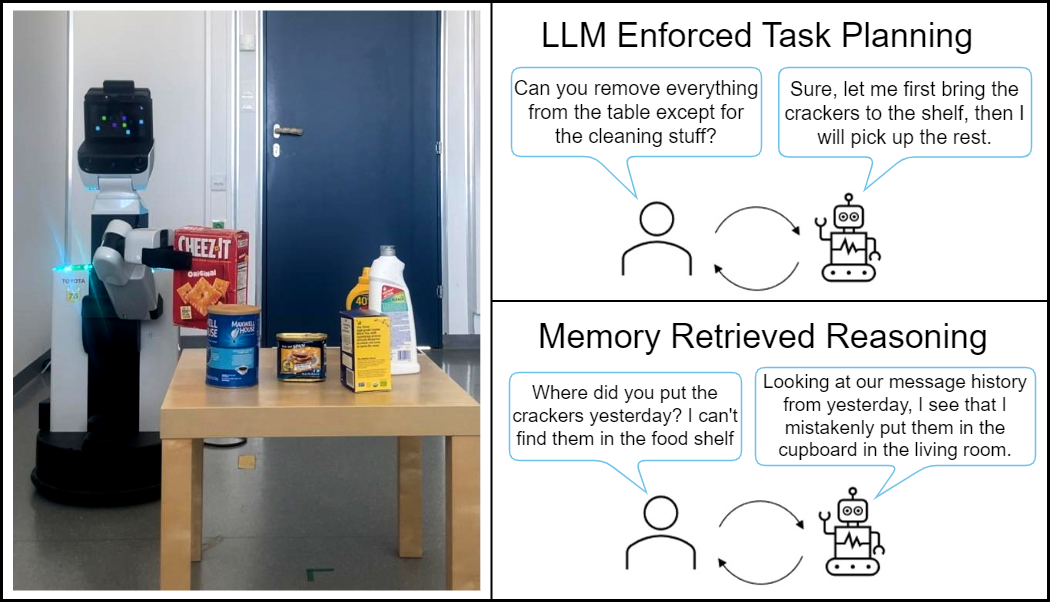}
    \caption{Our LLM-driven robotic system autonomously plans tasks and retrieves past interactions to improve object handling, illustrated by LLM-enforced task planning and memory-retrieved reasoning in a household setting.}
    \label{fig:enter-label}
\end{figure}

In this work, we present an embodied robotic system with an LLM-driven agent-orchestration architecture, where specialized software agents collaborate to address long-horizon household tasks.
Recent advances in Large Language Models (LLMs)~\cite{openai_gpt-4_2024, dubey2024llama, qwen_qwen25_2025, team2024gemma, guo2025deepseek} have improved systems real-world understanding, enabling common-sense reasoning in human language and making them accessible to researchers.
These advances combined with in-context learning \cite{wei_chain--thought_2022} enable flexible embodied task planning by decomposing high-level commands, such as \textit{"clear the dining table"}, into actionable steps based on detected objects~\cite{brohan_as_2023, huang_inner_2022, vemprala2024chatgpt, liang_code_2023, singh_progprompt_2023}.
By integrating Grounded Segment Anything Model (Grounded SAM)~\cite{ren2024grounded} and LLaMa3.2-Vision \cite{dubey2024llama}, our system creates grounded task plans. Unlike most other works, we address long-term operations by maintaining action and environment records, utilizing Retrieval-Augmented Generation (RAG) for efficient memory retrieval. Our approach enables the robot to autonomously organize and retrieve objects, interpret complex tasks, and provide updates on object locations, all while ensuring privacy through the use of offline LLMs and avoiding explicit model training. To illustrate the systems interaction, Fig.~\ref{fig:enter-label} shows an example of our system in action.


In summary, we present the following key contributions:
\begin{itemize}
    \item A long-horizon task planner for household tasks leveraging in-context learning and offline LLMs.
    \item Use of RAG for efficient memory retrieval and object tracking.
    \item A modular agent-orchestration system that improves robustness and modularity.
    \item Evaluation of the system's performance in three real-world household scenarios.
\end{itemize}

This paper is structured as follows: 
Section~\ref{sec:2_related_work} reviews related work in the areas of task planning and memory mechanisms. Section~\ref{sec:3_methodology} details the proposed system architecture. Section~\ref{sec:4_experiments} describes the experimental setup and household scenarios. Section~\ref{sec:5_results_and_discussion} presents the results. Finally, Section~\ref{sec:6_conclusion} concludes the paper and outlines directions for future work.

\section{Related Work}
\label{sec:2_related_work}
In this section we discuss related work for action and task planning, as well as memory and knowledge base.

\subsection{Action and Task Planning} 
Recent advancements in prompt engineering have improved the problem-solving capabilities of LLMs ~\cite{wei_chain--thought_2022, zhou_least--most_2022}, enabling the generation of structured plans without fine-tuning. Consequently, modern agent architectures leverage LLMs to dynamically react to execution failures \cite{yao_react_2023, huang_inner_2022} and expand their context by retrieval~\cite{lewis_retrieval-augmented_2020} or external tools \cite{schick_toolformer_2023, ruan2023tptu}.
However, LLMs lack an inherent understanding of a robot's physical abilities and real-world constraints. \textit{SayCan}~\cite{brohan_as_2023} addresses this by integrating value functions of pre-trained robotic skills to ensure feasibility, whereas Huang et al.~\cite{huang_language_2022} leverage LLMs to match high-level plans with low-level actions through semantic mapping.
Some works treat LLMs as programmers rather than direct decision-makers: \textit{Code-as-Policies}~\cite{liang_code_2023} and \textit{ProgPrompt} \cite{singh_progprompt_2023} allow LLMs to generate structured code for robotic executions, enhancing flexibility but adding an execution layer.

Pallagani et al.~\cite{pallagani_prospects_2024} found that LLMs perform better as translators of natural language into structured plans rather than generating plans from scratch. This ensures feasible actions based on predefined world models \cite{silver_generalized_2024, liu_delta_2024}.
These approaches are particularly effective in highly controlled environments, but present challenges when applied to open-ended, dynamic household settings.
Our work, instead, embraces flexible, dynamic task planning with in-context learning like shown in \cite{vemprala2024chatgpt}.
The approaches named, while effective for short-horizon tasks, do not track object positions over time. For long-horizon tasks that involve real-world dynamic conditions, a combination of task planning and a memory mechanism is required.

\subsection{Memory and Knowledge Base}
Long-horizon tasks require robust memory mechanisms. While LLM context windows keep expanding~\cite{team2024gemma}, using excessively large contexts in robotics is computationally inefficient.
Instead, long-term memory retrieval, accessed only when needed, is a more viable solution.
RAG~\cite{lewis_retrieval-augmented_2020} provides an efficient mechanism for narrowing context by querying a vast dataset and retrieving only relevant information.
Additionally, scene graphs, used in approaches like SayPlan~\cite{rana2023sayplan} and DELTA~\cite{liu_delta_2024}, offer structured memory that improves action verification and contextual reasoning. However, in unstructured and constantly changing environments, maintaining these graphs becomes challenging due to the need for complex automatic mechanisms or manual curation.

Our work explores the feasibility of a lightweight, fully natural language-driven approach using RAG as a memory mechanism. Inspired by ReMEmbR~\cite{anwar_remembr_2024}, our system incorporates temporal elements into the retrieval process, ensuring the robot tracks long-term changes in its environment.
While using language-based memory retrieval introduces potential for increased errors compared to structured models like scene graphs, we aim to evaluate how well purely language-based memory retrieval performs in practical, dynamic household scenarios. This approach offers flexibility, adaptability, and reduces the need for explicit world modelling, making it more suitable for real-world applications.

\section{Methodology}
\label{sec:3_methodology}

\begin{figure}[h]
    \centerline{\includegraphics[width=0.5\textwidth]{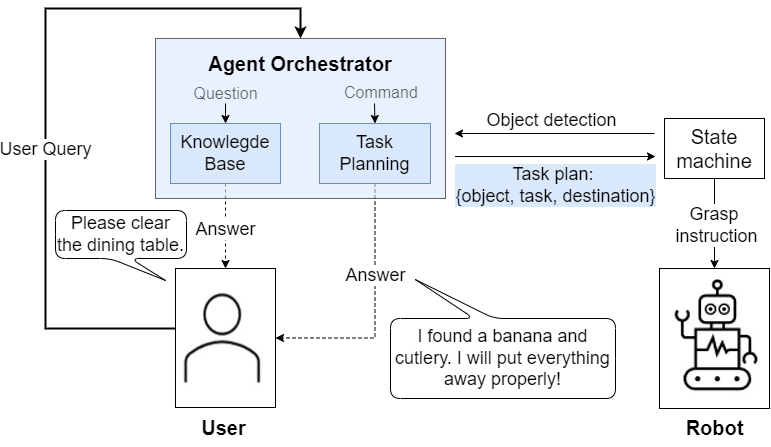}}
    \caption{The full pipeline, integrating long-horizon task planning. Newly introduced components are highlighted in blue.}
    \label{fig:full_pipeline}
\end{figure}

Our system, coordinated by an agent-orchestration framework, combines task planning with RAG~\cite{lewis_retrieval-augmented_2020}. This chapter explains the individual components and their interaction.

\label{sec:3_1_pipeline}

Fig.~\ref{fig:full_pipeline} illustrates the overall pipeline. The focus of this work is the agent-orchestration system, which processes object detection and user requests to create a robot task plan. In the system, each agent uses an LLM with a specialized role. The task planning agent additionally is prompted with a chain-of-thought technique \cite{wei_chain--thought_2022}.

\begin{figure}[h]
    \centering
    \includegraphics[width=1.0\linewidth]{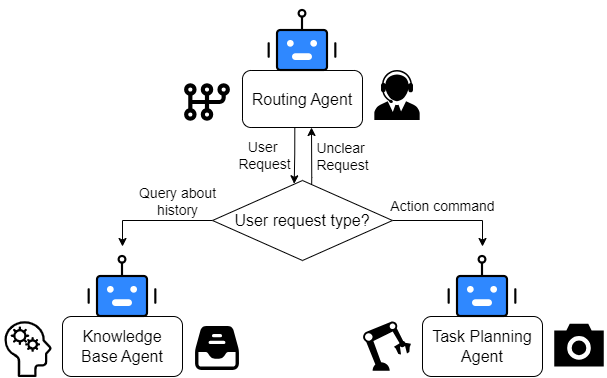}
    \caption{The agent-orchestration architecture}
    \label{fig:multi-agent}
\end{figure}

The system architecture of the agent orchestrator, illustrated in Fig.~\ref{fig:multi-agent}, consists of:
\begin{enumerate}
    \item A \textbf{routing agent}, responsible for analyzing incoming user requests.
    \item A \textbf{task planning agent}, handling commands that require the robot to perform actions.
    \item A \textbf{knowledge base agent}, processing follow-up questions about previously handled objects.
\end{enumerate}

When a user request arrives, the routing agent first analyzes it to determine its nature. The request is then categorized into one of three types: \\
\begin{enumerate}
    \item \textbf{Action command:} If the robot is asked to perform an action, it is forwarded to the task planning agent.
    \item \textbf{Query about history:} If it concerns previously handled objects, it is directed to the knowledge base agent.
    \item \textbf{Unclear request:} If the request doesn’t fit either category, clarification is requested before proceeding.
\end{enumerate}

\subsection{Task Planning Agent}
\label{sec:3_2_task_planning_agent}

The task planning agent receives frequent environmental updates via camera perception, encoded as a list of single objects. Grounded SAM~\cite{ren2024grounded} enables text-driven object detection and segmentation for the pipeline, while Vision Language Models (VLMs) generate natural language descriptions of the environment. Although VLMs alone can extract the object list for the LLM, Grounded SAM is essential for precise segmentation, which is critical for grasping tasks. 
Using the object list, the LLM processes the user request -- which can be both expressed in high-level or low-level terms -- and formulates tasks that best fulfill the command. The generated answer has to include a JSON string for an action following this structure:

\begin{enumerate}
    \item \textbf{Objects involved} in the task.
    \item \textbf{The destination} for placement tasks.
\end{enumerate}

After the action is determined, the grasping process is initiated.
We use the segmentation from Grounded SAM and the camera intrinsics to crop the depth image and project the depth crop to a 3D pointcloud of the respective object.
To estimate a grasp approach vector, we feed the cropped object point cloud to Control-GraspNet~\cite{sundermeyer2021contactgraspnet}, a pre-trained grasp estimator.

\subsection{Knowledge Base Agent}
\label{sec:3_3_knowledge_base_agent}

\begin{figure}[h]
    \centerline{\includegraphics[width=0.5\textwidth]{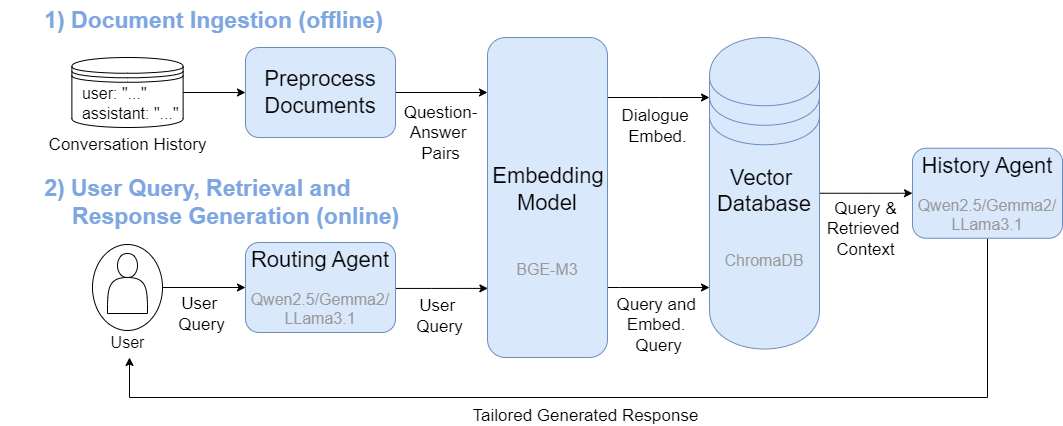}}
    \caption{RAG workflow for long-term question answering: Relevant past actions are retrieved from dialogue history, and the LLM generates responses based on the retrieved context.}
    \label{fig_rag_workflow}
\end{figure}

The knowledge base agent is used for user inquiries regarding past robot actions, such as object locations or task completion status. These queries require access to long-term memory, for which RAG has proven most effective, as discussed in Section~\ref{sec:2_related_work}.
Fig~\ref{fig_rag_workflow} illustrates the RAG workflow, comprising two key steps:

\begin{enumerate}
    \item \textbf{Document Ingestion:} Input data, such as conversation history, is preprocessed, split into smaller chunks (each representing a question-answer pair), and converted into high-dimensional vectors using an embedding model. These embeddings are then stored in a vector database for efficient retrieval.
    \item \textbf{User Query, Retrieval, and Response Generation:} User queries are embedded using the same model and are matched against the stored vectors to retrieve the most relevant context. This context is then provided to the LLM, which generates a response tailored to the user’s query.
\end{enumerate}

To enable chronological reasoning, essential for tracking object movements over time, we augment RAG with a time stamp for each question-answer pair.

\section{Experiments}
\label{sec:4_experiments}
To evaluate our system, we conduct experiments addressing the three key challenges from Chapter \ref{sec:1_introduction}: (1) flexible task planning in dynamic household environments, (2) long-term memory usage, and (3) modular agent coordination. Specifically, we assess the system’s ability to create grounded task plans, answer questions based on prior interactions, and route tasks to the appropriate agent.

\subsection{Experimental Setup}
\label{sec:4_1_experimental_setup}

This study evaluates an agent-orchestration system for symbolic task planning and follow-up questions via a knowledge base. To ensure a thorough evaluation, we consider three distinct phases:

\begin{enumerate}
    \item \textbf{Task Planning Performance} – The symbolic task planning output is assessed independently, measuring accuracy of object assignment to their destinations.
    \item \textbf{Knowledge Base Reliability} – The system’s ability to reason about past actions (with and without RAG) is tested by asking about the system's current status, such as locations of previously moved items.
    \item \textbf{Routing Reliability} – Measures the accuracy of the routing agent in directing queries to the appropriate agent (Task Planning, History, or itself).
\end{enumerate}

To isolate the performance of the specialized agents, agent handoff is not considered in the evaluation of 1) and 2).

\subsection{Algorithmic Framework}
The frameworks and models used are shown in gray in Fig.~\ref{fig_rag_workflow}. To enable efficient collaboration among agents, we use \textit{OpenAI Swarm}~\cite{openai_swarm}, a lightweight framework for agent orchestration and task delegation. We evaluate the performance of \textit{Qwen2.5-32b}~\cite{qwen_qwen25_2025}, \textit{Gemma2-27b}~\cite{team2024gemma}, and \textit{LLaMa3.1-8b}~\cite{dubey2024llama}, selected for their open-source availability and ability to run locally on 16GB GPU RAM. For RAG, we employ \textit{ChromaDB}~\cite{chroma}, a vector database optimized for fast lookups, combined with the embedding model \textit{BGE-M3}.

\subsection{Task Scenarios}
\label{sec:4.3_task_scenarios}
\begin{figure}[h]
    \centering
    \includegraphics[width=1.0\linewidth]{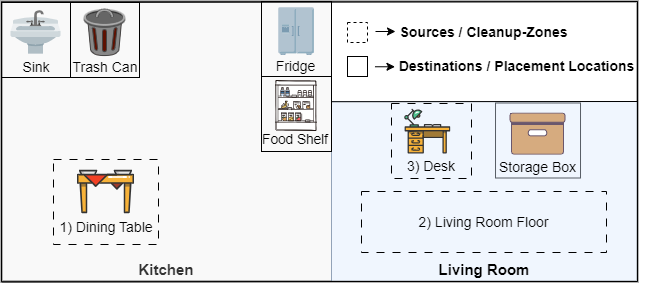}
    \caption{The artificial household environment used in the experiment.}
    \label{fig:environment_map}
\end{figure}

\begin{figure*}[t]
    \centering
    \begin{subcaptionbox}{\textbf{Scenario 1:} Dining Table Cleanup \\ \textbf{Object list from VLM:} \textit{Plate, Fork, Spoon, Salt shaker, Glass, Frying pan, Spatula, Chair, Table top, Pepper grinder.} \textbf{Command:} \textit{I just finished dinner, please clear the dining table. }}[.335\textwidth]
        {\includegraphics[width=\linewidth]{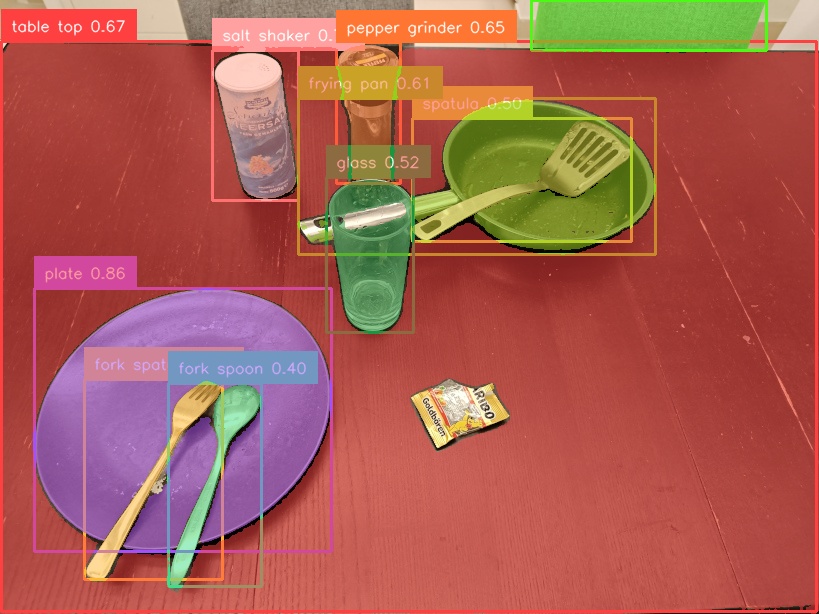}}
    \end{subcaptionbox}
    \hfill
    \begin{subcaptionbox}{\textbf{Scenario 2:} Living Room Cleanup \\ \textbf{Object list from VLM:}  \textit{A table, A couch, A brush, Scissors, Pen, Book, Salt packet, Jacket, Markers.}
\textbf{Command:} \textit{Please hand me the brush and tidy up the rest of the living room.}}[.3\textwidth]
        {\includegraphics[width=\linewidth, height=5cm, keepaspectratio]{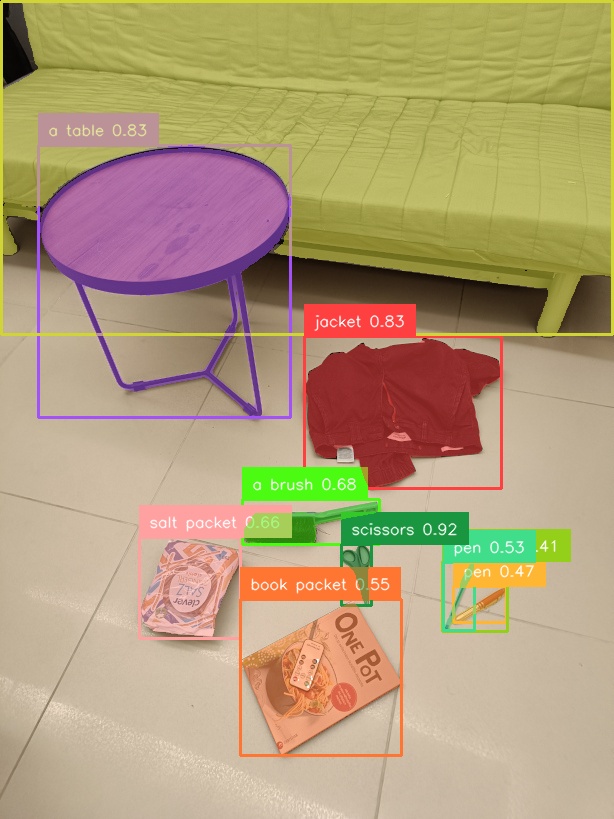}}
    \end{subcaptionbox}
    \hfill
    \begin{subcaptionbox}{\textbf{Scenario 3:} Desk Organization \\ \textbf{Object list from VLM:} \textit{Desk, Computer Monitor, Laptop, Mouse, Plate, Crumbs, Lemon, Cup, Glass of water, Bag of chips, Piece of paper, Potted plant, Cord, Wooden desk, White wall.} \textbf{Command:} \textit{Please clear my desk, leaving only the essentials for work.}}[.335\textwidth]
        {\includegraphics[width=\linewidth]{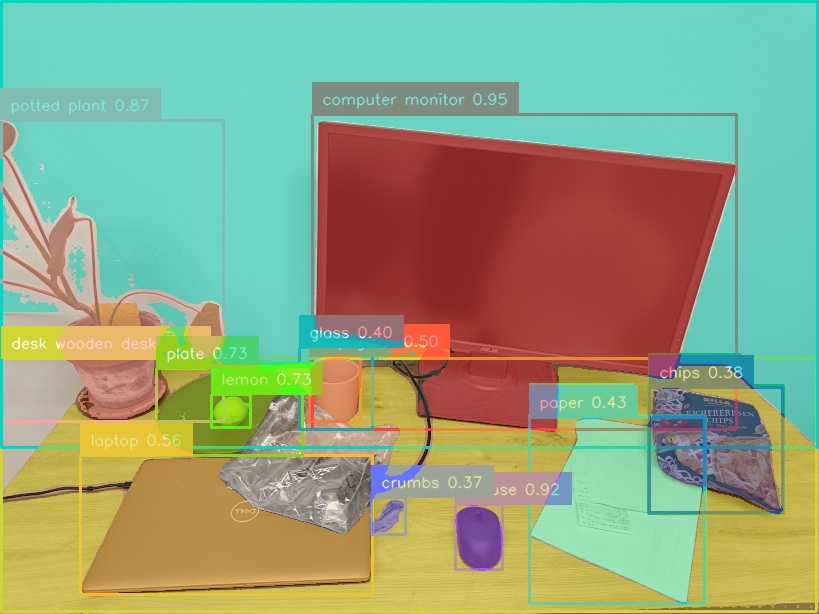}}
    \end{subcaptionbox}
    \caption{The three scenarios used for task planning. For each scenario we have extracted an object list using the Vision-Language Model \textit{LLaMa3.2-Vision}. This list is used as input for Grounded SAM~\cite{ren2024grounded} to perform segmentation.}
    \label{fig:seg_images}

\end{figure*}

The experiment is conducted in an artificial household environment, where objects must be assigned to correct destinations based on high-level commands. To evaluate task planning, we define three scenarios (see Fig. \ref{fig:seg_images}) that share five predefined placement locations, while each uses a different cleanup zone. Fig.~\ref{fig:environment_map} shows a visual representation of the environment. These locations reflect common-sense knowledge typically understood by LLMs. To ensure clarity, the agent receives explicit definitions for each destination:

\begin{itemize}
    \item \textbf{Sink} – For items that need washing.
    \item \textbf{Trash Can} – For disposable or inedible items.
    \item \textbf{Fridge} – For perishable food.
    \item \textbf{Food Shelf} – For non-perishable food items.
    \item \textbf{Storage Box} – For general storage.
\end{itemize}

\vspace{0.2cm}

Fig.~\ref{fig:seg_images} shows the object list extracted from a captured image of each task scenario using \textit{LLaMa3.2-Vision} along with the user queries and the segmentation results from Grounded SAM.

After execution of all scenarios, the knowledge base agent is asked four distinct folow-up questions targeting different aspects of retrieval and reasoning:
\begin{itemize}
    \item \textbf{Error Detection}: "Where is the jacket that was in the living room? I thought you put it in the storage box, but I can’t find it there."
    \item \textbf{Hallucination}: "Where did you put the laptop? It’s not on the desk anymore."
    \item \textbf{Food Availability}: "I am hungry. Is there any food left from earlier?"
    \item \textbf{Trash Status}: "How many objects are in the trash can?"
\end{itemize}

To better reflect real-world applications, we extend the conversation dialogue with additional question-answer pairs containing actions.
Furthermore, deliberate errors are introduced into the task plans, where the agent provides the user a different location than the one forwarded to the state machine.
This allows us to evaluate how well the knowledge base handles inaccuracies.
Beyond evaluating the specialized agents in isolated setups, we assess how effectively the routing agent delegates tasks to the appropriate specialized agent. Specifically, we test:

\begin{itemize}
    \item \textbf{Task Planning Queries}: The three high-level commands from the task planning scenarios (see Fig.~\ref{fig:seg_images}) and an additional low-level request ("Can I have a banana?") 
    \item \textbf{Knowledge Base Queries}: The four follow-up questions from the knowledge base scenario.
\end{itemize}

\subsection{Evaluation Methodology}
\label{sec:4.4_evaluation_methods}
The evaluation of the agent-orchestration system's components is based on the task scenarios and follow-up questions defined in Section \ref{sec:4.3_task_scenarios}.
Task planning performance is evaluated by testing each model on the three task scenarios, with each scenario executed five times per model. Accuracy is measured at the object level as the percentage of correctly assigned tasks. 
A task is deemed correct if it satisfies the following criteria:
\begin{itemize}
    \item \textbf{Valid JSON format}
    \item \textbf{Correct destination assignment}
    \item \textbf{Stationary Object Exclusion} (ensuring no task is assigned to items that should remain in place)
\end{itemize}
The final accuracy score represents the percentage of objects for which tasks were correctly assigned, including the implicit "no task" assignment for stationary objects (e.g., table).

The knowledge base is evaluated using four follow-up questions, each tested five times per model. Unlike the task planning agent, the knowledge base agent does not require a strict output format. It is assessed based on factual correctness, measured as the percentage of correct answers. For queries expecting multiple objects as an answer (e.g., "Which objects are in the trash?"), accuracy is based on the percentage of correctly identified objects.

The routing agent's ability to correctly assign tasks is evaluated by processing queries from the task planning scenarios and history-based questions, along with one additional query, five times per model. The final metric is quantified as the percentage of correctly assigned tasks. Gemma2, which does not support tool calling, is excluded from this test.

\section{Results and Discussion}
\label{sec:5_results_and_discussion}
This section presents the experimental results for task planning, knowledge base and agent routing.

\subsection{Task Planning}

\begin{table*}[htbp]
    \centering
    \renewcommand{\arraystretch}{1.2}

    \begin{tabular}{|l|cc|cc|cc|cc|}
        \hline
        \multirow{2}{*}{\textbf{Model}} & 
        \multicolumn{2}{c|}{\textbf{Dining Table}} & 
        \multicolumn{2}{c|}{\textbf{Living Room}} & 
        \multicolumn{2}{c|}{\textbf{Desk Organization}} & 
        \multicolumn{2}{c|}{\textbf{Total Accuracy (\%)}} \\
        \cline{2-9} 
        & \textbf{Strict (\%)} & \textbf{Lenient (\%)} & 
        \textbf{Strict (\%)} & \textbf{Lenient (\%)} & 
        \textbf{Strict (\%)} & \textbf{Lenient (\%)} & 
        \textbf{Strict (\%)} & \textbf{Lenient (\%)} \\
        \hline
        LLaMa3.1-8B   & \textbf{68.0} & 78.0 & 40.0 & 40.0 & 61.3 & 65.3 & 56.4 & 61.1 \\
        Gemma2-27B    & 58.0 & 68.0 & 68.9 & 68.9 & 68.0 & 69.3 & 65.0 & 68.7 \\
        Qwen2.5-32B   & 64.0 & \textbf{80.0} & \textbf{88.9} & \textbf{88.9} & \textbf{78.7} & \textbf{84.0} & \textbf{77.2} & \textbf{84.3} \\

        \hline
    \end{tabular}
    
    \vspace{1mm}
    \caption{Task Planning Accuracy Across Different LLMs. \textbf{Strict (\%)}: Percentage of objects correctly placed according to the intended plan. \textbf{Lenient (\%)}: Percentage of objects placed differently than expected, but with reasonable alternative placements based on user preferences.}
    \label{tab:eval_task_planning}
    
\end{table*}

We introduce a \textit{lenient} evaluation metric (cf. Table~\ref{tab:eval_task_planning}), where reasonable alternative placements based on user preferences are counted as correct.
The strictly correct placements, following the intended plan as prompted to the LLM, are presented under the \textit{strict} metric in Table~\ref{tab:eval_task_planning}.

Table~\ref{tab:eval_task_planning} shows that \textit{Qwen} consistently outperforms the other models in nearly all scenarios. \textit{LLaMA} performs notably worse in the living room scenario, with the lowest strict accuracy (40.0\%). \textit{Gemma2} falls between the two, showing higher accuracy than \textit{LLaMA} but lower than \textit{Qwen}.


\subsection{Knowledge Base}
\begin{table*}[htbp]
    \centering
    \renewcommand{\arraystretch}{1.2}
    \begin{tabular}{|l|l|cccc|c|}
        \hline
        \multirow{2}{*}{\textbf{Method}} & \multirow{2}{*}{\textbf{Model}} & \multicolumn{4}{c|}{\textbf{Response Validity (\%)}} & \multirow{2}{*}{\textbf{Total Validity (\%)}} \\
        \cline{3-6} 
        & & \textbf{Err. Detection} & \textbf{Hallucination} & \textbf{Food Avail.} & \textbf{Trash Status} & \\
        \hline
        \multirow{3}{*}{\textbf{Without RAG (Ablation Study)}} & LLaMa3.1-8B  & 20.0 & 80.0 & 70.0 & 65.0 & 58.8 \\
        & Gemma2-27B    & 0.0 & 80.0 & 10.0 & 60.0 & 37.5 \\
        & Qwen2.5-32B   & 0.0 & 80.0 & 60.0 & \textbf{75.0} & 53.75 \\
        \hline
        \multirow{3}{*}{\textbf{With RAG}} & LLaMa3.1-8B  & 40.0 & \textbf{100.0} & \textbf{90.0} & 55.0 & 71.25 \\
        & Gemma2-27B    & 80.0 & \textbf{100.0} & 40.0 & 60.0 & 70.0\\
        & Qwen2.5-32B   & \textbf{100.0} & \textbf{100.0} & \textbf{90.0} & \textbf{75.0} & \textbf{91.3}\\
        \hline
    \end{tabular}
    
    \vspace{1mm}
    \caption{Knowledge Base Response Accuracy Across Different LLMs. \textbf{Used Embedding Model for RAG}: \textit{BGE-M3}. \textbf{No. of question-answer pairs retrieved by RAG}: \textit{5} \\}
    \label{tab:knowledge_base_evaluation}

\end{table*}

The integration of RAG notably enhances the accuracy of the knowledge base's responses, even in medium-term interactions consisting of 21 question-answer pairs with approximately 4000 tokens. \textit{Qwen} achieves the highest validity (91.3\%) with RAG (cf. Table~\ref{tab:knowledge_base_evaluation}), highlighting the potential of retrieval-augmented approaches for maintaining consistency over longer interactions. 

\subsection{Agent Routing}

\begin{table*}[htbp]
    \centering
    \renewcommand{\arraystretch}{1.2}
    \begin{tabular}{|l|c|c|c|}
        \hline
        \textbf{Model} & \textbf{Task Planning Queries (\%)} & \textbf{Knowledge Base Queries (\%)} & \textbf{Total Success Rate (\%)} \\
        \hline
        LLaMa3.1-8B & 85.0 & \textbf{100.0} & \textbf{92.5} \\
        Qwen2.5-32B & \textbf{95.0} & 85.0  & 90.0 \\
        \hline
    \end{tabular}    
    \caption{Routing Success Rate Across Different LLMs}
    \label{tab:routing_accuracy}

\end{table*}

In task delegation, \textit{LLaMA} exhibits the highest routing accuracy (92.5\%), despite its weaker reasoning abilities (cf. Table~\ref{tab:routing_accuracy}). Its structured approach to tool-calling ensures stable performance. In contrast, \textit{Qwen}, while superior in contextual understanding, occasionally produces incorrect structured outputs, leading to execution failures.

\subsection{Summary}
Our findings highlight the potential of lightweight, open-source LLMs for memory-augmented long-horizon task planning. A combination of \textit{LLaMA} (routing) and \textit{Qwen} (specialized agents) achieves the best balance between structured execution and high-level reasoning.

Evaluating task execution remains challenging due to subjective human preferences, emphasizing the need for user studies.
Furthermore, integrating Vision-Language Models (VLMs) into the agent orchestrator -- rather than only using them for object lists -- could enhance robustness. Embedding contextual information into the latent space reduces command dependency and improves autonomy.

RAG improves factual consistency in knowledge retrieval but struggles with repeated object interactions and long histories, making full-history queries impractical.
Scene graphs, as proposed by Liu et al.~\cite{liu_delta_2024}, present a promising alternative for efficient and robust knowledge integration.

While task delegation via the routing agent was mostly successful, certain models occasionally produced invalid structured outputs, leading to execution failures. To increase robustness, future work should explore schema validation and adaptive retry mechanisms that can automatically mitigate such issues.

In summary, open-source LLMs prove viable for long-horizon task planning. However, addressing key challenges -- refining evaluation metrics, improving long-term robustness, and integrating multimodal perception -- remains essential for achieving reliable household robotics.

\section{Conclusion}
\label{sec:6_conclusion}
This work presents a prototype of an agent-orchestration system for household robots, utilizing local, lightweight open-source LLMs to translate high-level user commands into structured task plans for tidy-up scenarios. Memory-augmented task planning enables follow-up queries about past actions, improving user interaction and assisting in locating misplaced objects. Our evaluation shows strong task planning, routing, and knowledge retrieval. with Qwen2.5 excelling in reasoning-heavy tasks and LLaMA3.1 providing a more efficient routing solution.
However, RAG-based retrieval for general tasks remains a challenge, particularly for implicit queries where relevant information is not always found. Addressing these limitations is key to improving long-term reasoning and knowledge access. 

Future work will focus on robust storage solutions, improved knowledge representations, broader user studies with structured datasets for evaluating and benchmarking existing approaches. Enhancing communication and tool usage in agent-orchestration will be crucial for greater adaptability and autonomy in household robotics.

\addtolength{\textheight}{-12cm}   

\section*{ACKNOWLEDGMENT}
This research is supported by the EU program EC Horizon 2020 for Research and Innovation under grant agreement No. 101017089, project TraceBot, and the Austrian Science Fund (FWF), under project No. I 6114, iChores.

{\small
\bibliographystyle{IEEEtranS}
\bibliography{bibliography}
}

\end{document}